\documentclass[journal]{IEEEtran}
\usepackage{spconf,amsmath,graphicx}
\usepackage{amsmath,subfigure,epsfig,bbding}
\usepackage{graphicx}
\usepackage{epstopdf}
\usepackage[usenames]{color}
\usepackage{cite}
\usepackage{enumerate}
\usepackage{mathrsfs}
\usepackage{amsfonts}
\usepackage{multirow}
\usepackage[boxed,commentsnumbered,ruled]{algorithm2e}
\usepackage{colortbl,array,booktabs}
\usepackage{changepage}
\usepackage{framed}
\usepackage{array}
\usepackage{booktabs} 
\usepackage{graphicx}
\usepackage{xcolor,stfloats}

\usepackage{lipsum}
\usepackage[noend]{algpseudocode}

\title{Query-Efficient Adversarial Attack Based on Latin Hypercube Sampling}
\name{Dan Wang, Jiayu Lin, and Yuan-Gen Wang$^{\ast}$}
\address{School of Computer Science and Cyber Engineering, Guangzhou University, Guangzhou, China}

\begin{document}
	\maketitle
	\begin{abstract}
		In order to be applicable in real-world scenario, Boundary Attacks (BAs) were proposed and ensured one hundred percent attack success rate with only decision information. However, existing BA methods craft adversarial examples by leveraging a simple random sampling (SRS) to estimate the gradient, consuming a large number of model queries. To overcome the drawback of SRS, this paper proposes a Latin Hypercube Sampling based Boundary Attack (LHS-BA) to save query budget. Compared with SRS, LHS has better uniformity under the same limited number of random samples. Therefore, the average on these random samples is closer to the true gradient than that estimated by SRS. Various experiments are conducted on benchmark datasets including MNIST, CIFAR, and ImageNet-1K. Experimental results demonstrate the superiority of the proposed LHS-BA over the state-of-the-art BA methods in terms of query efficiency. The source codes are publicly available at https://github.com/GZHU-DVL/LHS-BA.
	\end{abstract}
	
	\vspace{-2mm}
	\section{Introduction}
	\label{sec:intro}
	Deep learning models have achieved impressive results in various domains, such as image classification, malware detection, speech recognition and medicine. With the use-case continual development, research on the vulnerabilities and threats posed by deep learning is gaining attention. In order to test the anti-interference and robustness of deep learning models, researchers proposed the concept of adversarial examples \cite{Szegedy2014, Goodfellow2015, Dong2017}. That is, attackers can generate adversarial examples which are visually similar to the original images, but mislead the deep neural network to giving the wrong outputs. For instance, traffic signs may be modified with small stickers, leading to incorrect classifications. As done in \cite{Eykholt2018}, stop signs are recognized as speed limit signs. Similarly, facial recognition systems can be easily fooled by people with a pair of colored glass \cite{Sharif2016}.
	
	Early methods \cite{Goodfellow2015, Szegedy2014} for generating perturbed images  operate mainly in a white-box setting. The attacker has access to all information about the model. However, this white-box setup is clearly unrealistic since the model parameters are not always exposed to the attackers. Current attack methods prefer the black-box settings where only the output scores and hard labels are available. Some recent works \cite{Narodytska2016, Chen2017} have focused on the score-based attacks. They exploit the output probabilities to generate adversarial examples, while require a large number of queries to the target model. Most existing black-box attack methods do not take into account query cost, especially for commercial models. Indeed, some commercial models provide users with only the final decisions, even without the output probabilities. Therefore, the most practical setting is one where only the hard label can be observed, also known as the decision-based black-box setting. Under this setting, Brendel et al. \cite{Brendel2017} developed a Boundary Attack (BA). This method crafts adversarial examples through a simple rejection sampling and achieves comparable performance with state-of-the-art white-box attacks such as C\&W attack \cite{Carlini2017}. As an important category of adversarial attacks, BA is highly relevant to real-world applications and significantly important for measuring the robustness of models. Inspired by BA \cite{Brendel2017}, several methods \cite{Ilyas2018, Cheng2018} were proposed to perform the decision-based black-box attacks. However, they lack of efficiency due to requiring a great deal of queries or  getting a relatively large perturbation with a limited query budget.	
	 
	To address this problem, we design a Latin Hypercube Sampling based Boundary Attack (LHS-BA). Compared with the existing BA methods, our LHS-BA can successfully attack different models on individual dataset with much fewer queries. The main contributions of this paper are two-fold: (1) The Latin Hypercube Sampling (LHS) is adopted to generate the random samples around the adversarial example. Compared to the simple random sampling (SRS) method, LHS has better symmetry with the same limited number of random samples. Therefore, the components in the non-gradient direction can be better canceled out, and the true gradient can be estimated more accurately. (2) Extensive experiments show that our LHS-BA can dramatically reduce the queries to the target model and achieve better performance than the state-of-the-art BA methods.
	
	\begin{figure}
		\centering    
		\includegraphics[scale=.28]{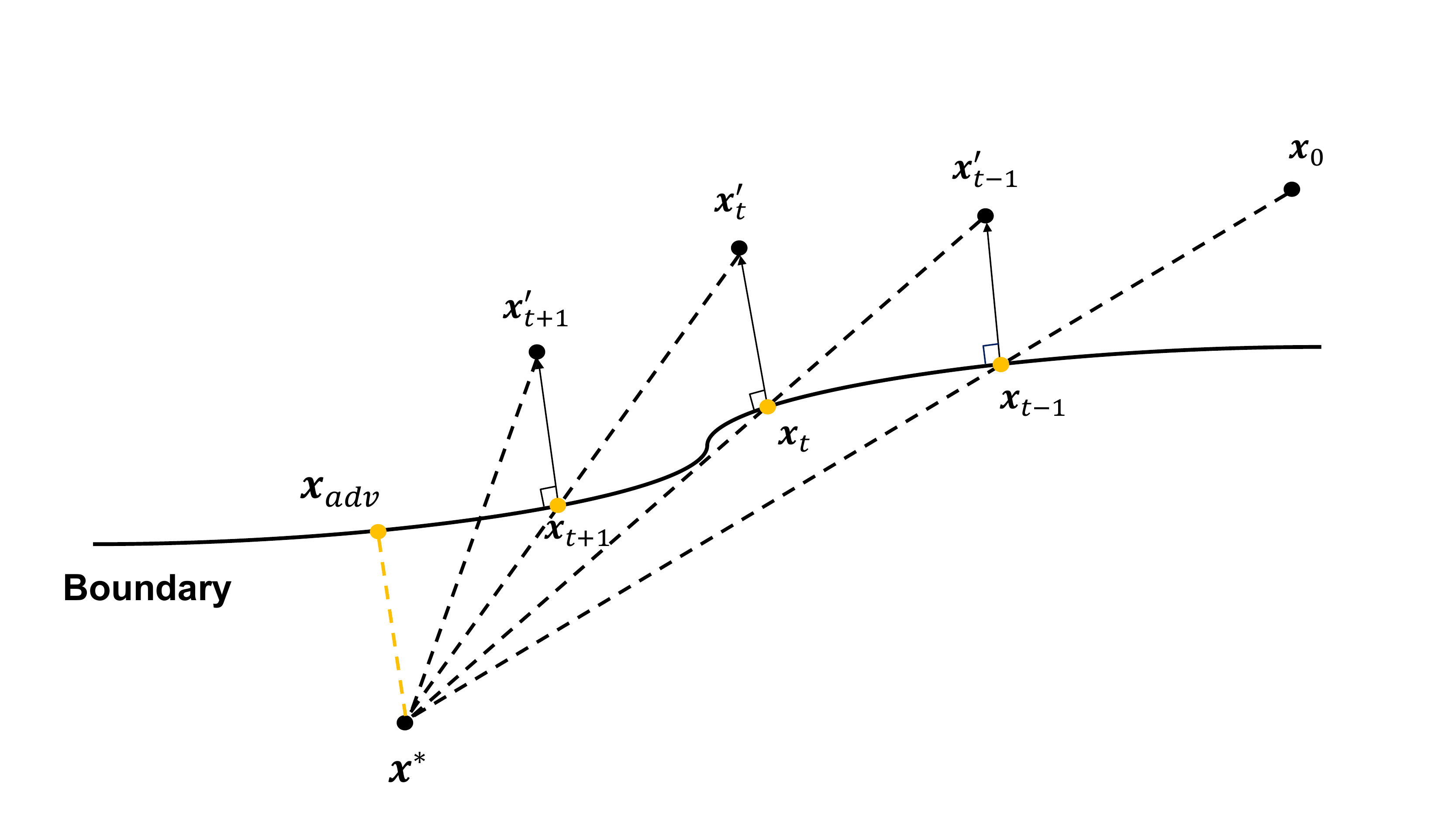}
		\vspace{-9mm}
		\caption{Three steps to craft an adversarial example with LHS-BA: (1) The binary search algorithm is used to search the starting point located at the boundary. (2) Using $M$ random vectors $\boldsymbol{n}_{1}, \ldots, \boldsymbol{n}_{M}$ to estimate the direction of the gradient. (3) Search the smallest distortion adversarial example in the estimated gradient direction.} \label{fig:innwd}
		\vspace{-3mm}
	\end{figure}

	\section{Problem Definition}
	\label{sec:format}
	\textbf{\emph}  We consider a trained model $ f: 
	\boldsymbol{x} \rightarrow c$, where $\boldsymbol{x}\in \mathbb{R } ^{n} $ is an input normalized image and $c\in[k]$ is the final decision of the model with $k$ classes (e.g. the top-1 classification label). Given an input image $\boldsymbol{x}$,  the model may output a $k$-dimensional vector $F(\boldsymbol{x})$ which represents the probability distribution over all classes. Assume that the label of an input image $\boldsymbol{x}^{*}$ is $c^{*}:=\arg \max _{c \in[k]} F_{c}(\boldsymbol{x}^{*})$. Untargeted attack aims to change the classifier decision $c^{*}$, and the goal of targeted attack is to make the model misclassified into a certain pre-specified class $c^{+}$. Define the function $J$ and the indicator $C$ by
	\begin{equation}\label{}
\vspace{-1mm}
J_{\boldsymbol{x}^{*}}(\boldsymbol{x}):=\left\{\begin{array}{ll}
\max \limits_{c \neq c^{*}} F_{c}(\boldsymbol{x})-F_{c^{*}}(\boldsymbol{x}) &\text {(Untargeted)} \\
F_{c^{+}}(\boldsymbol{x})-\max \limits_{c \neq c^{+}} F_{c}(\boldsymbol{x}) &\text {(Targeted)}
\end{array}\right.
\vspace{-3mm}
\end{equation}
	
	\begin{equation}
	C_{\boldsymbol{x}^{\ast}}\left(\boldsymbol{x}\right)=\textrm{sign}\left(J_{\boldsymbol{x}^{\ast}}\left(\boldsymbol{x}\right)\right)=\left\{
	\begin{array}{ll}
	1 & \textrm{if} \,\, J_{\boldsymbol{x}^{\ast}}\left(\boldsymbol{x}\right)>0 \\
	-1 & \textrm{otherwise}
	\end{array}\right.
	\vspace{-1mm}
	\end{equation}
	\noindent
	In decision-based black-box settings, the attacker computes a perturbation $\boldsymbol{n}$ to change the estimated label to an incorrect label without any knowledge about the target model. In the boundary attack, the value of $C$ (NOT $J$) is only able to get. Given a perturbed example $\boldsymbol{x}^{\ast}+\boldsymbol{n}$ which is sent to the  to model queries, $C_{\boldsymbol{x}^{*}}\left(\boldsymbol{x}^{\ast}+\boldsymbol{n}\right) = 1$ is defined as a successful attack. We craft adversarial examples by solving the following optimization problem,
	\begin{equation}\label{} 
	\begin{array}{ll}
	\min _{} & \mathcal{D}(\boldsymbol{x}^{*}, \boldsymbol{x}^{*}+\boldsymbol{n}) \\
	\text { s.t. } & C(\boldsymbol{x}^{*}+\boldsymbol{n}) = 1
	\end{array}
	\end{equation}
	where $\mathcal{D}(\cdot, \cdot)$ is an $l_{2}$ distance metric.
	
	\begin{figure}
		\centering
		\includegraphics[width=3.2in]{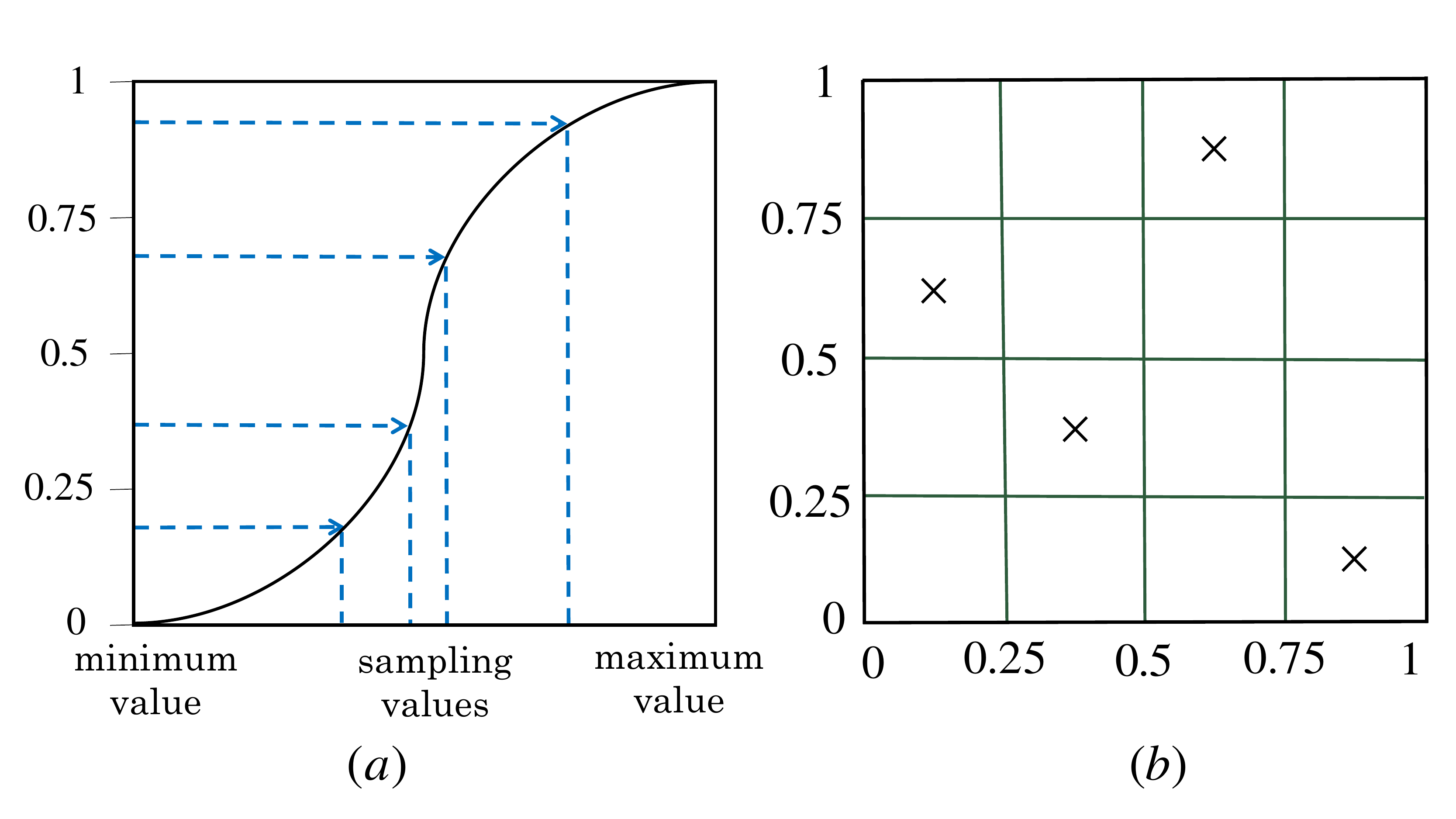}
		\vspace{-3mm}
		\caption{Illustration of Latin Hypercube Sampling of obtaining four samples.} \label{fig:innwd}
	\vspace{-2mm}
	\end{figure}
	
	\section{Proposed method}
	In this section we present the main idea of the Latin Hypercube Sampling (LHS) and the pipeline of the proposed LHS-BA. Fig. 1 provides an intuitive visualization of the proposed method and the detailed implementation is described in Algorithm 1. LHS-BA first initializes an adversarial example at decision boundary and then performs an iterative algorithm which consists of three steps: 1) Estimate the gradient direction, 2) Take a step forward in the estimated gradient direction, and 3) Project onto decision boundary. More details about each step are given below.
	
	\subsection{Latin Hypercube Sampling}	
	As a method for sampling from a given probability distribution, LHS \cite{McKay1979} is a form of stratified sampling. It can be used to sample the random numbers in a sample space where samples may be arbitrarily distributed. Furthermore, it is usually applied to high-dimensional variables and achieve good randomness. In uncertainty analysis, LHS usually requires fewer samples and converges faster than the Monte Carlo Simple Random Sampling (MCSRS) method.
	
	Suppose we would like to obtain $M$ samples from the normal distribution with zero mean and unit standard deviation.
	The idea behind one-dimensional LHS is easy to be understood. As shown in Fig. 3(a), LHS divides a given cumulative distribution function into $M$ identical intervals and randomly chooses one value from each cumulative distribution interval to obtain $M$ samples. The purpose is to split the total area under the probability density function into $M$ equal portions. It ensures that each interval will contain the same number of samples, thus producing good uniformity and symmetry. We can easily extend the one-dimensional LHS concept to two dimensions. As shown in Fig. 3(b), for a two-dimensional random vector, we can divide the cumulative distribution space of each dimension into $M$ identical intervals. Hence, we can obtain $M\times M$ cells. A square grid containing sample positions is a Latin Square if and only if there is only one sample in each row and each column. It is important to note that these two components must be independent to each other in order to obtain the samples we expect.

	\begin{figure}
		\centering	
		\includegraphics[scale=.25]{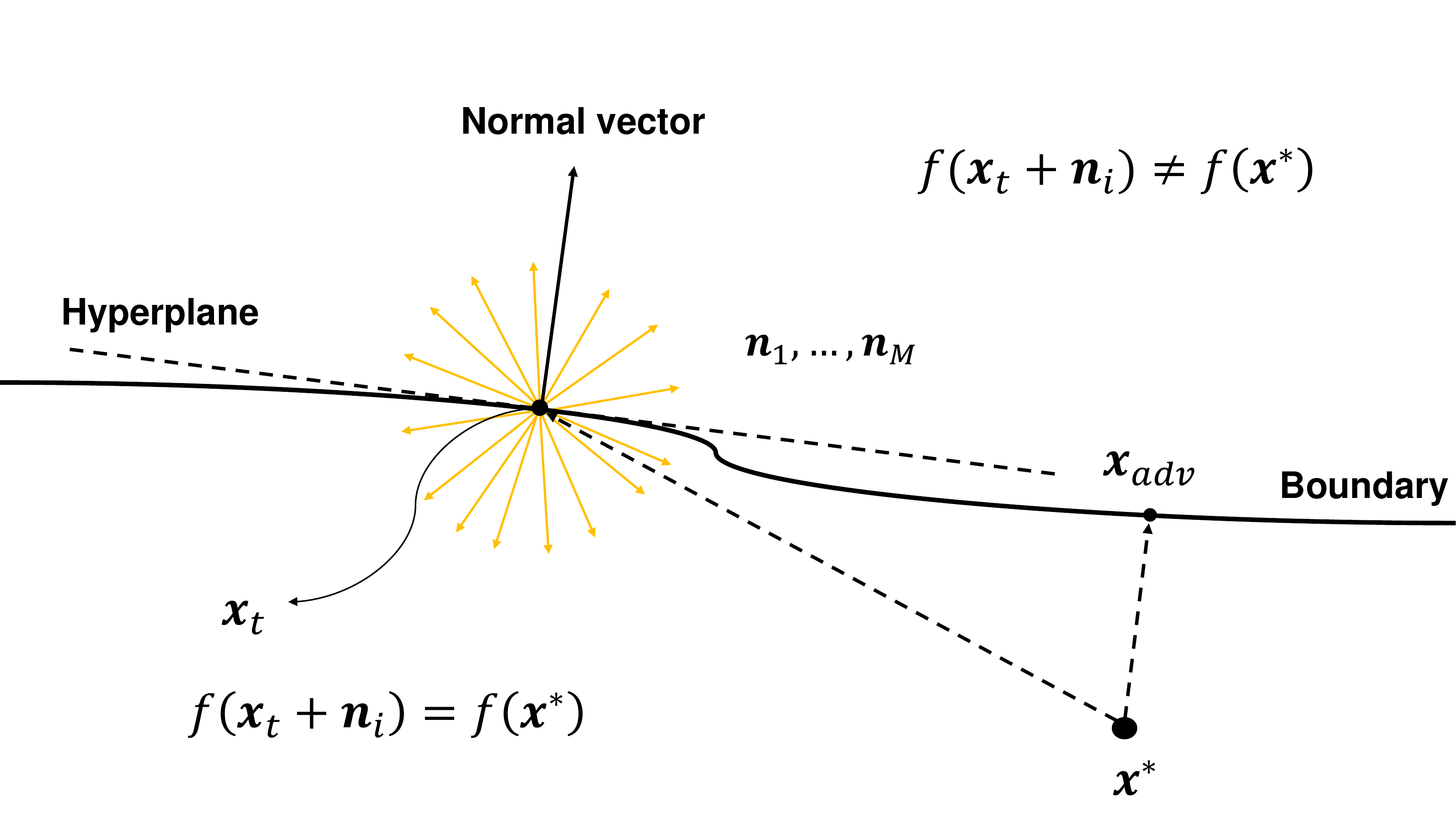}
		\vspace{-3mm}
		\caption{Overview of estimating gradient at decision boundary.} \label{fig:innwd}
		\vspace{2ex}
	\end{figure}
	
	Also, we can simply extend the idea of two-dimensional LHS into even more dimensions. The Latin Hypercube is the generalization of two-dimensional LHS to arbitrary dimensions, wherein each sample space is the only one in each axis-aligned hyperplane containing it. When the sample space of $N$ dimensions is sampled, the value range of each variable is divided into $M$ equal probability intervals. Then $M$ sample points are chosen to satisfy the Latin Hypercube requirement. LHS is particularly advantageous in dealing with high-dimensional data. We can obtain the random samples that reflects the true underlying distribution. Even though LHS is not completely random, it requires a smaller sampling number to achieve the same accuracy as the simple random sampling. In other words, the sample value of LHS can cover the entire distribution interval of the random variable. As sample standard deviation is smaller, the samples obtained by LHS are more evenly distributed.
	
	\vspace{-2.7mm}
	\subsection{Estimate gradient at decision boundary}
	In Algorithm 1, we first initialize $\boldsymbol{x}_{0}$. For untargeted attack, a noise example drawn from the uniform distribution is used as an initial adversarial example. For targeted attack, the initial image is randomly selected from the dataset. Given an adversarial example at the decision boundary $\boldsymbol{x}_{t}$ at the $t$-th step, we can estimate the gradient of $J(\boldsymbol{x}_{t})$ by sending queries to the target model,
	\begin{equation}\label{}
	{\nabla J}(\boldsymbol{x}_{t}):=\frac{1}{M} \sum_{i=1}^{M} C_{\boldsymbol{x}^{*}}\left(\boldsymbol{x}_{t}+\delta \boldsymbol{n}_{i}\right) \boldsymbol{n}_{i}
	\end{equation}
	where $\left\{\boldsymbol{n}_{i}\right\}$ are $M$ random vectors, and $\delta$ is a small positive number. The estimation process is also illustrated in Fig. 3. It has been shown that the more symmetrically the sampled noise vectors are distributed, the better the components in the other directions than the gradient direction will cancel out \cite{Liu2019}. 
	With a finite number of sampling noise vectors, the random vectors sampled by LHS not only have good uniformity but also good symmetry, which makes the gradient estimation more accurate.
	\begin{algorithm}
		\caption{Latin Hypercube Sampling based Boundary Attack}
		{\bf Input:} \hspace*{0.02in}
		Classifier $C$, original image $\boldsymbol{x}^{*}$, sampling number $M$, total number of iterations $T$, positive parameter $\delta$, step size $\epsilon$, stopping threshold of binary search $\theta$.\\
	{\bf Output:}
		An adversarial example. \\
		Initialize $\boldsymbol{x}_{0}$ which lies on the decision boundary.\\
		\For {$t = 0$ to $T-1$}{	
			
			Sample $M$ unit vectors $\boldsymbol{n}_{1}, \ldots, \boldsymbol{n}_{M}$ via Latin Hypercube Sampling.\\
			Estimate ${\nabla J}(\boldsymbol{x}_{t})$ with the rule defined in Eq. (4).\\
			Set $\boldsymbol{x}^{\prime} = \boldsymbol{x}_{t} + \epsilon \cdot \frac{{\nabla J}}{\|{\nabla J}\|_{2}}$.\\
			Perform binary search for $\boldsymbol{x}_{t+1}$ = BinSearch($\boldsymbol{x}^{\prime}, \boldsymbol{x}^{*}, C$).}
		Output final adversarial example $\boldsymbol{x}_{t+1}$.
		
	\end{algorithm}
	\setlength{\intextsep}{5pt}
	\setlength{\textfloatsep}{10pt}

	\subsection{Take a step forward} 
	Once the gradient is estimated, the $\boldsymbol{x}_{t}$ will moved one step forward towards the gradient direction:
	\begin{equation}\label{}
	\boldsymbol{x}^{\prime} = \boldsymbol{x}_{t} + \epsilon \cdot \frac{{\nabla J}}{\|{\nabla J}\|_{2}}
	\end{equation}
	where $\epsilon$ is the size of perturbation at the $t$-th step. Note that $\boldsymbol{x}^{\prime}$ is at the opposite side of the boundary to $\boldsymbol{x}^{*}$. 
	\vspace{-4mm}
	\subsection{Project onto decision boundary}
	Since the gradient estimation works only at the boundary, we need to project $\boldsymbol{x}^{\prime}$ onto the boundary. The binary search algorithm is used to quickly find the approximate boundary between the adversarial area and the non-adversarial one. The adversarial example at next step is found until the distance of the adversarial example to the decision boundary is smaller than a given stopping threshold $\theta$. Hence, we move the adversarial image $\boldsymbol{x}^{\prime}$ towards the original image $\boldsymbol{x}^{*}$,
	\begin{equation}\label{}
	\boldsymbol{x}^{t+1} = \alpha_t \cdot \boldsymbol{x}^{*} + (1 - \alpha_t) \cdot \boldsymbol{x}^{\prime}
	\end{equation}
	where $\alpha_t$ is a changing positive parameter between 0 and 1 in order that $\boldsymbol{x}^{\prime}$ can be projected back to the decision boundary.
	
	\section{Experiments}
	\label{sec:format} In this section, we perform extensive experiments to demonstrate the effectiveness of the proposed LHS-BA. And we compare the performance of LHS-BA with several existing decision-based attacks on image classification tasks.
	
	\vspace{-3mm}
	\subsection{Experimental Settings}
	\textbf{\emph{1) Datasets.}} Four benchmark datasets are used for testing: MNIST, CIFAR-10 \cite{Krizhevsky2009}, CIFAR-100 \cite{Krizhevsky2009} and ImageNet-1K \cite{Deng2009}. MNIST contains 70K $28\times?28$ size handwritten digits images in the range [0, 9]. CIFAR-10 has 10 classes and 6K images per class. CIFAR-100 has 100 classes and 600 images per class. ImageNet-1K has 1,000 classes and images in ImageNet-1K are rescaled to $224\times 224\times 3$. For MNIST and CIFAR-10, we use 1,000 correctly classified test images, which are randomly drawn from the test dataset, and evenly distributed across all classes. For ImageNet-1K, 100 correctly classified test images are used, evenly distributed among 10 randomly selected classes. The number of random vectors ($M$) is set to 100 in the first attack, and then gradually increases according to $M = M \times (t+1)^{\frac{1}{5}}$, where $t$ denotes the $t$-th iteration. At the $t$-th iteration, we compute $\delta_{t} = \|\boldsymbol{x}_{t-1}-\boldsymbol{x}^{*}\|_{2} / m$ as probe step size in each gradient estimation,  where $m$ = 224$\times$224$\times$3 is the input dimension. Meanwhile, we set $\epsilon_{t} = \|\boldsymbol{x}_{t-1}-\boldsymbol{x}^{*}\|_{2} / \sqrt{t}$ as perturbation step size in moving along estimated gradient direction. Stopping threshold of binary search is set to $\theta=m^{-\frac{3}{2}}$. Total number of iterations is set to $T=64$.
	
	\textbf{\emph{2) Target models.}} 
	In order to verify the robustness of attack methods against different network architectures, we evaluate proposed method on some popular models. For MNIST, we use a CNN architecture which contains four convolutional layers and three fully connected layers and shows a test error rate of 0.81$\%$. It is widely adopted as a victim model in many related works. For CIFAR-10 and CIFAR-100, we consider training a 20-layer ResNet-20\cite{He2016} with a test error rate of 7.56$\%$ and a 121-layer DenseNet\cite{Huang2017} with a test error rate of 8.40$\%$. For ImageNet-1K, we adopt a pre-trained 50-layer ResNet-50\cite{He2016}. We clip the perturbed example into [0,1] by default for all experiments.
	
	\textbf{\emph{3) Compared methods.}}
	We compare the performance of the proposed LHS-BA method with two state-of-the-art methods for decision-based black-box attacks, including the Boundary Attack method \cite{Brendel2017} and HopSkipJumpAttack method \cite{Chen2020}. We utilize the implementation of the two algorithms with the suggested hyperparameters from the publicly available source code online.
	\begin{table}
	\caption{Mean $\l_{2}$-norm distortion for performing untargeted attacks with different query budgets (1K, 2K, and 3K).}
	\vspace{2ex}
	\centering
	\begin{minipage}{1.0\linewidth}\Huge
		\centering
		\label{table:snetv2}
		\resizebox{1\textwidth}{!}{
			\begin{tabular}{cccccc}
				\hline  { Dataset } & \text { Victim Model }& \text { Method } &  1K&  5K &20K\\

				\hline & & \text { Boundary Attack\cite{Brendel2017} } & 10.950 & 9.628 & 2.539\\

				\text { MNIST } & \text { CNN } & \text { HopSkipJumpAttack \cite{Chen2020}} & 3.049 & 1.925 & 1.713\\

				& & \text {Proposed LHS-BA} & $\mathbf{2.858}$ & $\mathbf{1.784}$ & $\mathbf{1.560}$  \\

				\hline & & \text { Boundary Attack \cite{Brendel2017}} & 2.859 & 2.640 & 0.335\\

				& \text { ResNet-20 } & \text { HopSkipJumpAttack\cite{Chen2020} } &  {0.817}   & 0.271 & 0.164 \\
				& & \text {Proposed LHS-BA} & $\mathbf{0.745}$ & $\mathbf{0.235}$ & $\mathbf{0.158}$ \\[-2ex]
				\text { CIFAR-10 } \\ \cline{2-6}& & \text { Boundary Attack \cite{Brendel2017}} & 2.843 & 2.209 &2.310\\
				& \text { DenseNet } & \text { HopSkipJumpAttack\cite{Chen2020} } & 0.690 & 0.318 &0.192 \\
				& & \text {Proposed LHS-BA} & $\mathbf{0.548}$ & $\mathbf{0.297}$ &$\mathbf{0.190}$  \\
				
				\hline & & \text { Boundary Attack \cite{Brendel2017}} & 1.965 & 1.319 & 0.135\\
				& \text { ResNet-20 } & \text { HopSkipJumpAttack\cite{Chen2020} } &  {0.267} & 0.138 & 0.064 \\
				& & \text {Proposed LHS-BA} & $\mathbf{0.255}$ & $\mathbf{0.132}$ & $\mathbf{0.063}$ \\[-2ex]
				\text { CIFAR-100 } \\ \cline{2-6}& & \text { Boundary Attack \cite{Brendel2017}} & 1.870 & 1.543 &0.230\\
				& \text { DenseNet } & \text { HopSkipJumpAttack\cite{Chen2020} } &  0.368 & 0.219 &0.084 \\
				& & \text {Proposed LHS-BA} & $\mathbf{0.341}$ & $\mathbf{0.205}$ &$\mathbf{0.082}$ \\		
				
				\hline& & \text { Boundary Attack\cite{Brendel2017} } & 77.083 & 34.592 &6.749 \\
				\text { ImageNet-1K} & \text { ResNet-50 } & \text { HopSkipJumpAttack \cite{Chen2020}} & 40.745 & 11.716 &5.129 \\
				& & \text {Proposed LHS-BA} & $\mathbf{33.598}$ & $\mathbf{9.830}$  &$\mathbf{4.954}$  \\
				
				\hline
			\end{tabular}
		}		
	\end{minipage}
	\vspace{1ex}
\end{table}\label{table1}

\begin{table}[t]
	\caption{Mean $\l_{2}$-norm distortion for performing targeted attacks with different query budgets (1K, 2K, and 3K).}
	\vspace{2ex}
	\centering
	\begin{minipage}{1.0\linewidth}\Huge
		\centering
		
		\label{table:snetv2}
		\resizebox{1\textwidth}{!}{
			\begin{tabular}{cccccc}
				\hline  { Dataset } & \text { Victim Model }& \text { Method } &  1K&  5K &20K\\
				\hline & & \text { Boundary Attack\cite{Brendel2017} } & 10.179 & 9.970 & 2.692 \\
				\text { MNIST } & \text { CNN } & \text { HopSkipJumpAttack \cite{Chen2020}} & 3.281 & 2.376 & 2.054\\
				& & \text {Proposed LHS-BA} & $\mathbf{2.758}$ & $\mathbf{1.970}$ & $\mathbf{1.696}$  \\
				
				\hline & & \text { Boundary Attack \cite{Brendel2017}} & 9.625 & 8.649 & 8.464\\
				& \text { ResNet-20 } & \text { HopSkipJumpAttack\cite{Chen2020} } &  {4.226} & 0.654 & 0.277 \\
				& & \text {Proposed LHS-BA} & $\mathbf{3.066}$ & $\mathbf{0.509}$ & $\mathbf{0.256}$\\[-2ex]
				
				\text { CIFAR-10 }  \\ \cline{2-6}& & \text { Boundary Attack \cite{Brendel2017}} & 8.390  & 5.699  &0.427\\
				& \text { DenseNet } & \text { HopSkipJumpAttack\cite{Chen2020} } &  {1.857} & 0.462 &0.209 \\
				& & \text {Proposed LHS-BA} & $\mathbf{1.692}$ & $\mathbf{0.438}$ & $\mathbf{0.197}$  \\
				
				\hline & & \text { Boundary Attack \cite{Brendel2017}} & 9.795 & 7.829 & 1.793\\
				& \text { ResNet-20 } & \text { HopSkipJumpAttack\cite{Chen2020} } &  {6.528} &1.037 & 0.368 \\
				& & \text {Proposed LHS-BA} & $\mathbf{6.305}$ & $\mathbf{0.890}$ & $\mathbf{0.357}$ \\[-2ex]			
				\text { CIFAR-100 }  \\ \cline{2-6}& & \text { Boundary Attack \cite{Brendel2017}} & 9.038 & 6.922 &1.206\\
				& \text { DenseNet } & \text { HopSkipJumpAttack\cite{Chen2020} } &  {5.260} & 0.858 &0.274 \\
				& & \text {Proposed LHS-BA} & $\mathbf{5.087}$ & $\mathbf{0.797}$ &$\mathbf{0.270}$  \\
				\hline& & \text { Boundary Attack\cite{Brendel2017} } & 87.739 & 45.690 &10.595 \\
				\text { ImageNet-1K} & \text { ResNet-50 } & \text { HopSkipJumpAttack \cite{Chen2020}} & 68.802 & 39.806 &9.811 \\
				& & \text {Proposed LHS-BA} & $\mathbf{62.957}$ & $\mathbf{35.073}$  &$\mathbf{8.186}$  \\
				\hline
			\end{tabular}
		}		
	\end{minipage}
	\vspace{1ex}
\end{table}\label{table2} 
	
	\vspace{-3mm}
	\subsection{Experimental Results}
	The experimental results for the untargeted and targeted attacks are shown in Table I and Table II, respectively. We can clearly see from Tables I and II that the proposed LHS-BA performs best in terms of distortion reduction, especially in the early stages of an attack. It is important when only one low query budget is allowed. This is because the proposed method introduces LHS, which makes the estimated gradient more accurate and achieves better efficient query. We can also find that when attacking ImageNet-1K, due to the high-dimensional images, the three compared attack methods still need tens of thousands of model queries to reduce the $\l_{2}$-norm distortion of adversarial examples to be lower than 10, crafting the imperceptible adversarial examples. 
	
	Besides, we observe from Tables I and II that under the same queries, datasets, and models conditions, the untargeted attack can obtain lower-distorted adversarial examples than the targeted attack. This shows the fact that the targeted attack is actually harder than the untargeted attack. However, a counterintuitive phenomenon appears that the distortion of adversarial examples for CNN model on MNIST is  larger than that on other three datasets, which was also observed in previous work \cite{Chen2020}. This might be explained that it is more difficult to cheat the model on the simple tasks. We speculate that the query budget is highly related to the smoothness of input dimensions and decision boundaries.

	

	\section{Conclusion}
	\label{sec:ref}
	In this paper, we have presented a new  adversarial boundary attack method based on Latin Hypercube Sampling, termed LHS-BA. We use the binary search algorithm to search the initial attack position, and then estimate the gradient of the decision boundary by observing the network decision results. Finally, we project the adversarial example to the boundary again to facilitate the next gradient estimation. Throughout the attack, LHS makes the estimated gradient direction more accurate, which improves the query efficiency. We have studied LHS-BA through extensive experiments and confirmed its superior performance against the existing  black-box attacks in terms of the query efficiency, while without compromising in attack success rate.

	
\end{document}